\documentclass[cameraready]{Interspeech}

\usepackage{comment}
\usepackage{bm}
\usepackage{multirow}
\usepackage{xcolor}
\usepackage{graphicx}  
\usepackage{makecell}  
\usepackage{float}
\usepackage[T1]{fontenc}
\usepackage[utf8]{inputenc}

\title{Synthetic Speech, Real Signal: Paralinguistic Preservation and \\ Cross-Lingual Augmentation via Voice Cloning}
\author[affiliation={1},correspondingauthor]{Roseline}{Polle}
\author[affiliation={1}]{Owen}{Parsons}
\author[affiliation={1}]{George}{Fairs} 
\author[affiliation={1}]{Luis M}{San Martin Fernandez}
\author[affiliation={2}]{Cole}{Looney}
\author[affiliation={3}]{Xiaoliang}{Wu}
\author[affiliation={1}]{Alexandra L}{Georgescu}
\author[affiliation={1}]{Stefano}{Goria}

\address{
    $^1$ thymia, UK \\
    $^2$ The University of Edinburgh, UK \\
    $^3$ University of Southampton, UK
}
\email{roseline@thymia.ai}

\begin{document}
\maketitle

\begin{abstract}
Synthetic data augmentation in speech is common practice for linguistic tasks like ASR, but has seen far less work for paralinguistic ones, especially clinical tasks where labelled data is expensive and some patient groups are underrepresented. Voice cloning is one such augmentation approach, but is typically evaluated on speech intelligibility (WER) or speaker similarity (SS) rather than on downstream performance, and it remains unclear whether these preserve the paralinguistic signal such tasks depend on. We benchmark eight voice cloning models on five paralinguistic tasks across public and clinical datasets, showing most preserve signal with modest degradation. We then clone English clinical speech into Japanese and find that training on cloned data outperforms raw cross-lingual transfer for depression and anxiety detection on real Japanese speech, suggesting voice cloning is a promising direction for augmenting clinical speech data in low-resource languages.
\end{abstract}

\section{Introduction}
\label{sec:intro}

Synthetic speech augmentation is well explored for linguistic tasks such as Automatic Speech Recognition (ASR) ~\cite{Czy_nikiewicz_2024, laptev2020, bartelds2023, ogun2025}, but is less explored in paralinguistic tasks, such as emotion recognition and clinical biomarker detection~\cite{Norbury_2026, cummins2018speech}, where the need is greater due to labelled data being more expensive and harder to obtain.
Existing efforts focus on emotion recognition~\cite{shahid2023, ibrahim2024, chatziagapi2019, malik2023}, but these efforts target a single task and none adopts a voice cloning approach.

Voice cloning itself is typically evaluated on intelligibility (WER), speaker similarity (SS), and naturalness (MOS)~\cite{azzuni2025}, with broader paralinguistic preservation receiving only sporadic attention. ClonEval~\cite{christop2025cloneval} benchmarks cloning on emotional speech but measures only embedding similarity, and the VoicePrivacy Challenge~\cite{tomashenko2024voiceprivacy} uses emotion preservation as a metric for anonymisation rather than cloning fidelity.
No prior work directly evaluates whether cloned speech retains sufficient signal for downstream paralinguistic classification.

This gap is particularly pronounced in clinical speech. Although detection of depression and anxiety from speech has advanced rapidly~\cite{Norbury_2026, cummins2015review, low2020review}, nearly all large-scale labelled datasets are in English and not publicly available~\cite{Norbury_2026,kintsugi2024,sondehealth2024}. Voice cloning could help augment such scarce data while enabling privacy applications by decoupling what is said from how it is said, for instance when transcripts contain sensitive information. However, this approach is only viable if the relevant paralinguistic signal is preserved during the cloning process. 

We address three research questions: \textbf{RQ1.}~Do voice cloning models preserve paralinguistic information? \textbf{RQ2.}~Does this preservation 
extend across languages? \textbf{RQ3.}~Do standard cloning quality metrics predict paralinguistic preservation? 
To answer these, we benchmark eight  open-source models on five tasks (emotion, sentiment, sarcasm, accent, and depression/anxiety detection) across four public and one proprietary clinical dataset. For RQ2, we clone English clinical speech into Japanese and test whether models trained on cloned data can detect depression and anxiety in real Japanese speech.

\section{Method}
\label{sec:method}

\subsection{Voice Cloning Models}
\label{sec:models}

We evaluate eight open-source voice cloning models that cover autoregressive, flow-matching, and hybrid architectures (Table~\ref{tab:models}), selected for architectural diversity and reported quality.

\begin{table}[th]
  \caption{Voice cloning models evaluated. Only the models marked with $\dagger$ are used for RQ2 (cross-lingual).}
  \label{tab:models}
  \centering
  \resizebox{\columnwidth}{!}{%
  \begin{tabular}{ l l }
    \toprule
    \textbf{Model} & \textbf{Architecture} \\
    \midrule
    XTTS v2 & GPT-2 autoregressive + Perceiver \\
    Zonos & Mamba2 SSM + Transformer decoder \\
    E2-TTS & Flow-matching, character-level input \\
    F5-TTS & Flow-matching, ConvNeXt text encoder \\
    OpenAudio S1-mini$^*\dagger$ & Qwen3 LLM + online RLHF \\
    CosyVoice 2$^\dagger$ & Autoregressive + flow-matching \\
    CosyVoice 3$^\dagger$ & LLM + flow-matching, RL post-training \\
    MaskGCT$^\dagger$ & Masked generative codec transformer \\
    \bottomrule
    \multicolumn{2}{l}{\footnotesize *Hereafter referred to as OpenAudio.}
\end{tabular}}
\end{table}

\subsection{Datasets and Tasks}
\label{sec:datasets}

Table~\ref{tab:datasets} summarises the datasets and tasks used. For RQ1, we use four public datasets covering emotion, sentiment, sarcasm, and accent classification, as well as a proprietary English clinical speech corpus for depression and anxiety detection. For RQ2, the full proprietary corpus is cloned into Japanese and evaluated on a matched proprietary Japanese clinical speech corpus. We use a proprietary dataset for the clinical analyses as existing public alternatives are orders of magnitude smaller~\cite{gratch2014distress, tao2023androids,tasnim2022depac} and lack \emph{matched} non-English clinical speech.

\textbf{Proprietary dataset.\footnote{The English corpus was collected via Prolific under independent ethical review (ARMA, https://arma.ac.uk/; favourable outcome September 2, 2024) with written informed consent for voice-based mental health research. The Japanese corpus was collected via Rakuten Insight, with participants compensated through Rakuten credits (i.e. comparable to the original protocol).
Voice cloning was performed as an internal processing step, in accordance with UK GDPR Article 89 research provisions and with appropriate safeguards in place.}~\cite{Norbury_2026}} 
Participants completed three speech activities: \textit{paragraph} reading, a \textit{mood}-related question, and a \textit{general} question about their weekend, along with the PHQ-8 and GAD-7 questionnaires and a self-reported mental health history. A participant is labelled depressed/anxious if their PHQ-8/GAD-7 score is 10 or greater and they report a prior diagnosis.
\textbf{\textit{English (EN):}} 82,046 samples (${\sim}$228\,h in total) from 30,537 speakers (F/M=60/40\%; Mean age=37; White/Black/Asian/Mixed=70/15/7/5\%) across three activities (paragraph: 30,181; mood: 26,998; general: 24,867). For RQ1 we use a random ${\sim}$10k-user subset per activity; for RQ2 the full corpus is used.
\textbf{\textit{Japanese (JP):}} 14,159 samples (${\sim}$39\,h in total) from 6,253 speakers (F/M/unspecified=40/50/10\%; Mean age=$51$; Asian/Other/unspecified=85/5/10\%) collected under the same protocol (paragraph: 5,869; mood: 3,945; general: 4,345), used as the target-language test set for RQ2.

\begin{table}[th]
  \caption{Datasets and tasks.}
  \label{tab:datasets}
  \centering
  \resizebox{\columnwidth}{!}{%
  \begin{tabular}{ l c c c l }
    \toprule
    \textbf{Dataset} & \textbf{RQ} & \textbf{Samples} & \textbf{Speakers} & \textbf{Task(s)} \\
    \midrule
    IEMOCAP~\cite{busso2008iemocap} & 1 & 5,531  & 10  & Emotion [4 cls] \\
    MELD~\cite{poria2019meld} & 1  & 13,708 & 304 & Emotion [7 cls] / Sentiment [3 cls] \\
    MUSTARD~\cite{castro2019mustard} & 1 & 690    & 21  & Sarcasm [binary] \\
    VCTK~\cite{yamagishi2019vctk} & 1   & 29,650 & 73  & Accent [3 cls] \\
    \midrule
    Proprietary (EN)~\cite{Norbury_2026} & 1,2 & 82,046 & 30,537 & Depression [binary] / Anxiety [binary] \\
    Proprietary (JP) & 2   & 14,159  & 6,253  & Depression [binary] / Anxiety [binary] \\
    \bottomrule
  \end{tabular}}
\end{table}

\subsection{Processing Pipeline}
\label{sec:pipeline}

Recordings are transcribed with Whisper medium~\cite{radford2023robust}, trimmed of leading/trailing silence, and truncated to 10s to balance capturing speaker information with reasonable computational cost, and to ensure compatibility across all models evaluated. Each segment then serves as reference audio for cloning under two text conditions:
\textbf{\textit{Repeat}:} the cloned segment reproduces the original transcript;
\textbf{\textit{Standard:}} all cloned speakers produce the same fixed passage, removing linguistic content and isolating paralinguistic signal.

For RQ2, we apply the \emph{repeat} condition cross-lingually: English transcripts are translated to Japanese using Qwen~3 235B~\cite{qwen3} via Amazon Bedrock, and the cloning model generates Japanese speech from the translated text using the original English recording as speaker reference. 
Since the paragraph activity already has all speakers read a fixed passage (from which the standard cloning text is drawn), repeat and standard conditions are nearly equivalent for this activity. For RQ2 we therefore omit standard-text cloning and report paragraph-only performance to disentangle semantic from paralinguistic signal.

\subsection{Experimental Setup}
\label{sec:setup}

\textbf{Features and classifier.}
We extract 1024-dimensional embeddings from WavLM Large~\cite{chen2022wavlm}, chosen for its strong paralinguistic performance~\cite{zhang2024paralbench}. We use a standard scaler and classify with Logistic Regression (L2, $C{=}0.001$, balanced class weights). All results are reported as AUC (one-vs-rest, macro-averaged for multi-class).
\textbf{Comparison to benchmarks.}
We report AUC differences with baselines in percentage points (pp).
We also define a \textit{preservation score} $P = (A_{\text{c}} - 0.5) / (A_{\text{r}} - 0.5)$, where $A_{\text{c}}$ and $A_{\text{r}}$ are AUC on cloned and real speech respectively, measuring the fraction of above-chance signal retained ($P{=}1$: perfect; $P{=}0$: chance). This metric is only defined where AUC is significantly above chance. When summarising Preservation scores, we report means across tasks for each model and medians across models for each task, the latter to limit the influence of outlier models.

\textbf{RQ1: Monolingual preservation.}
For each dataset we compare \textbf{Real} (train and test on originals) with \textbf{Cloned} (train and test on cloned segments) under the \emph{repeat} and \emph{standard} text conditions.  We perform stratified group 5-fold cross-validation (CV) with no speaker leakage, and test significance with a permutation test (AUC vs.\ chance) and a paired bootstrap test (cloned vs.\ Real, $10{,}000$ speaker-level resamples).

\textbf{RQ2: Cross-lingual transfer.}
Using the four cross-lingual models marked in Table~\ref{tab:models}, we compare three training conditions: \textbf{in-language} (JP Real, train and test on Japanese real speech), \textbf{raw cross-lingual} (Real EN, train on real English, test on Japanese real), and \textbf{cloned cross-lingual} (EN$\to$JP, train on English cloned into Japanese, test on Japanese real). 
The raw cross-lingual baseline (Real EN) is the most straightforward transfer strategy, helping isolate the effect of cloning, since our aim is not to establish state-of-the-art transfer.
For the cross-lingual conditions, we train on stratified subsets of N=$10{,}000$ English speakers (5 random seeds) and test on the full Japanese corpus (5 bootstrap resamples), for a total of 25 iterations. Significance vs.\ Real EN is assessed with a paired z-test. A scaling analysis varies the number of English training speakers from 50 to ${\sim}30{,}000$. The in-language condition serves as reference but uses a different protocol (on the full JP dataset) and is not directly comparable: we use 5-fold CV and bootstrap-resample speakers (50 iterations) to obtain stable confidence intervals.

\textbf{RQ3: Speaker similarity vs.\ downstream preservation.}
To investigate whether standard voice cloning metrics predict paralinguistic preservation, we focus on speaker similarity, as the other common metrics are less suitable for benchmarking: Word Error Rate (WER) depends on the choice of Automatic Speech Recognition (ASR) system, and Mean Opinion Score (MOS) is subjective. We compute the mean cosine similarity between real and cloned WavLM speaker embeddings for each model--dataset pair in RQ1, following the same protocol as ClonEval~\cite{christop2025cloneval}, and correlate these values with the corresponding AUC degradation (cloned vs.\ Real).

\section{Results}
\label{sec:results}

\subsection{RQ1: Paralinguistic Preservation}
\label{sec:results_rq1}

\begin{table}[h]
  \caption{AUC on public datasets under both cloning conditions.
  \textbf{Bold} = best, \underline{underline} = second best per column;
  $\dagger$~marks models not significantly different from Real
  ($p \geq 0.05$, bootstrap paired test).}
  \label{tab:rq1_public_combined}
  \centering
  \resizebox{\columnwidth}{!}{%
  \begin{tabular}{lcccccc}
    \toprule
    \textbf{Model} & \textbf{\makecell{IEMOCAP\\Emotion}} & \textbf{\makecell{MELD\\Emotion}} & \textbf{\makecell{MELD\\Sent.}} & \textbf{\makecell{MUSTARD\\Sarcasm}} & \textbf{\makecell{VCTK\\Accent}} & \textbf{Mean} \\
    \midrule
    Real & 0.896 & 0.696 & 0.712 & 0.661 & 0.957 & 0.785 \\
    \midrule
    \multicolumn{7}{l}{\textit{Repeat}} \\
    \midrule
    E2-TTS & 0.856 & 0.670 & \underline{0.686} & \textbf{0.690}$^{\dagger}$ & 0.938 & \textbf{0.768} \\
    OpenAudio & 0.833 & \textbf{0.681} & \textbf{0.688} & 0.680$^{\dagger}$ & \textbf{0.946} & \underline{0.766} \\
    MaskGCT & \textbf{0.869} & \underline{0.677} & 0.674 & 0.658$^{\dagger}$ & 0.938 & 0.763 \\
    CosyVoice 3 & \underline{0.867} & 0.674 & 0.683 & 0.637$^{\dagger}$ & 0.935 & 0.759 \\
    F5-TTS & 0.848 & 0.662 & 0.676 & \underline{0.688}$^{\dagger}$ & 0.921 & 0.759 \\
    CosyVoice 2 & 0.852 & 0.669 & 0.674 & 0.648$^{\dagger}$ & 0.925 & 0.754 \\
    Zonos & 0.813 & 0.660 & 0.657 & 0.628 & \underline{0.944}$^{\dagger}$ & 0.740 \\
    XTTS v2 & 0.822 & 0.655 & 0.657 & 0.639$^{\dagger}$ & 0.907 & 0.736 \\
    \midrule
    \multicolumn{7}{l}{\textit{Standard}} \\
    \midrule
    MaskGCT & \underline{0.844} & \underline{0.615} & \textbf{0.629} & 0.632$^{\dagger}$ & 0.873 & \textbf{0.719} \\
    E2-TTS & 0.819 & \textbf{0.621} & \underline{0.625} & \underline{0.655}$^{\dagger}$ & 0.851 & \underline{0.714} \\
    Zonos & 0.797 & 0.590 & 0.611 & 0.568 & \textbf{0.995} & 0.712 \\
    CosyVoice 2 & 0.827 & 0.605 & 0.615 & \textbf{0.667}$^{\dagger}$ & 0.834 & 0.710 \\
    CosyVoice 3 & \textbf{0.845} & 0.613 & 0.623 & 0.630$^{\dagger}$ & 0.822 & 0.706 \\
    F5-TTS & 0.805 & 0.613 & 0.616 & 0.618$^{\dagger}$ & 0.837 & 0.698 \\
    XTTS v2 & 0.790 & 0.582 & 0.595 & 0.623$^{\dagger}$ & \underline{0.881} & 0.694 \\
    OpenAudio & 0.793 & 0.601 & 0.608 & 0.615$^{\dagger}$ & 0.845 & 0.693 \\
    \bottomrule
  \end{tabular}}
\end{table}

\begin{figure*}[th]
  \centering
  \includegraphics[width=\textwidth]{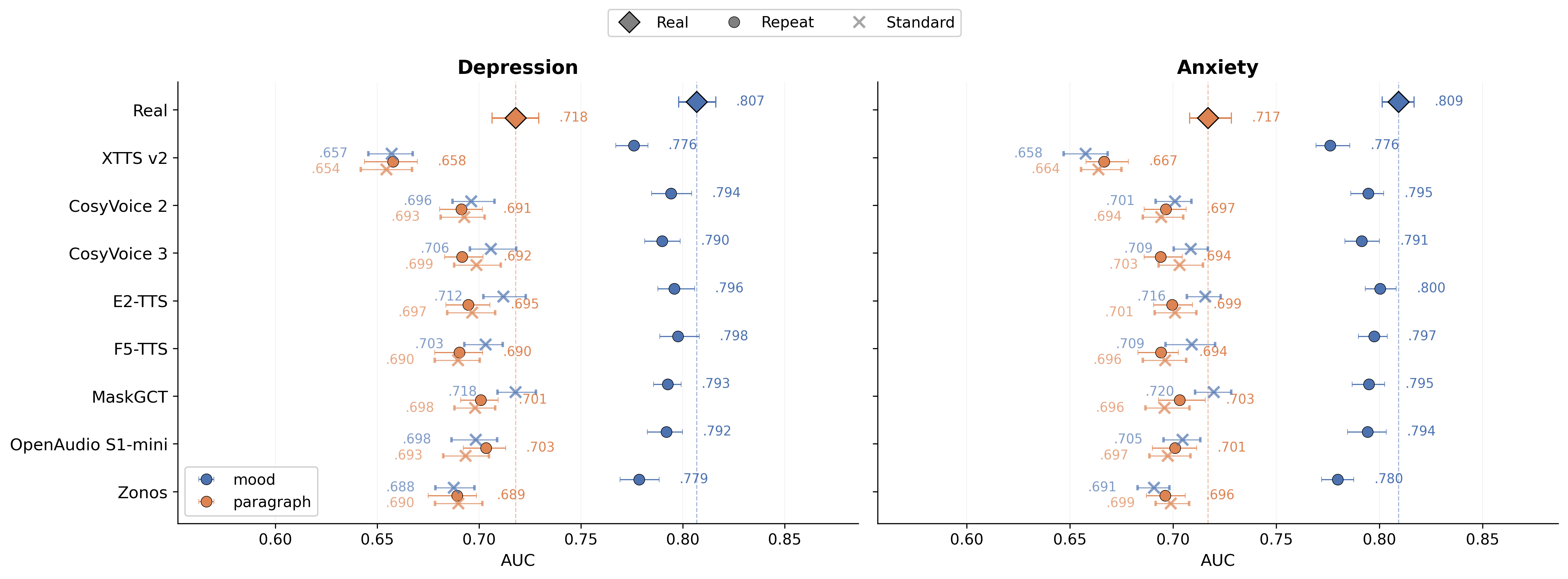}
  \caption{AUC on the proprietary clinical datasets (depression and anxiety)
  under both cloning conditions. Circles = Repeat (right annotations);
  \texttimes{} = Standard (left annotations).
  Colours denote speech activities (mood, paragraph); the general activity is omitted as results closely mirror paragraph.
  Dashed lines show Real AUC baselines.}
  \label{fig:rq1_mdc}
\end{figure*}

Table~\ref{tab:rq1_public_combined} and Figure~\ref{fig:rq1_mdc} report AUC scores for the public and clinical datasets respectively. All 176 model--task configurations achieve significantly above-chance performance (permutation test, $p{<}0.05$), confirming that paralinguistic signal survives cloning. While 161 show statistically significant degradation from Real (paired bootstrap, $p{<}0.05$), the drop is modest (median 3.2\,pp; median $P{=}0.87$).

\textbf{Repeat condition.}
On the public benchmarks (Table~\ref{tab:rq1_public_combined}), all eight models preserve the majority of paralinguistic signal, with mean degradation ranging from $-1.7$\,pp to $-4.9$\,pp (Preservation score $P{=}0.94$ to $0.83$; defined in Section~\ref{sec:setup}). Five models (E2-TTS, OpenAudio, MaskGCT, CosyVoice3, F5-TTS) retain over 90\% of the above-chance signal ($P{\geq}0.90$). Across tasks, accent classification is most robust (median $P{=}0.95$), while emotion (median $P{=}0.87$) and sentiment (median $P{=}0.82$) show more loss. On the clinical corpus (Figure~\ref{fig:rq1_mdc}), preservation is comparable ($P{=}0.93$--$0.94$ for the top models), with mood showing the highest retention (median $P{=}0.95$) and paragraph/general the most loss (median $P{=}0.89$). Depression and anxiety results are numerically similar due to high comorbidity (${\sim}85\%$ label overlap), consistent with known co-occurrence rates~\cite{Nc_Mg_2017,McGrath2020}.

\textbf{Standard condition.}
Under the standard condition, removing linguistic content reduces preservation substantially (median $P{=}0.75$ vs.\ $P{=}0.90$ for repeat). The drop is largest for tasks where content carries diagnostic signal: mood speech falls from median $P{=}0.95$ ($-1.5$\,pp) to $P{=}0.66$ ($-10.4$\,pp), the largest repeat--standard gap, while paragraph reading is virtually unchanged (median $P{=}0.89$--$0.90$ in both conditions, ${\sim}$$-2.1$\,pp). This confirms that paragraph performance reflects paralinguistic rather than linguistic signal, validating its use in RQ2 to disentangle the two (as specified in Section \ref{sec:pipeline}).
Even accent classification, the most robust task under repeat ($P{=}0.95$), drops to $P{=}0.76$ under the standard condition, but absolute performance remains high ($>0.8$ AUC). Zonos is a notable outlier: it achieves 0.995 AUC on VCTK, exceeding even the Real baseline (0.957), despite ranking in the bottom three models on all other standard-condition tasks.

\subsection{RQ2: Cross-Lingual Transfer}
\label{sec:results_rq2}

Given that paralinguistic signal survives monolingual cloning, we next explore whether this extends across languages. Of the six models with native Japanese support, we select the four strongest RQ1 performers for RQ2; XTTS~v2 and Zonos are excluded as the weakest under the repeat condition, while E2-TTS and F5-TTS do not support cross-lingual cloning.

Table~\ref{tab:rq2} reports cross-lingual AUC at 10{,}000 English training speakers. All four cloned conditions significantly outperform the raw
cross-lingual baseline ($p{<}.001$), with OpenAudio showing the largest gain for depression ($+3.3pp$) and CosyVoice3 for anxiety ($+4.0 pp$), followed by MaskGCT. A gap remains relative to the in-language Japanese reference, which trains on ${\sim}$4{,}000 labelled speakers using within-corpus cross-validation rather than cross-corpus transfer (Section~\ref{sec:setup}), and is therefore not directly comparable.

Cloning benefits even paragraph reading ($+2.3$\,pp depression, $+4.5$\,pp anxiety for CosyVoice3), whose diagnostic signal is largely content-independent (similar AUC degradation under both repeat and standard in RQ1). This suggests the advantage is not coming only from translating semantic content, but also from adapting the acoustic space to better match the target language.

\begin{table}[t]
  \caption{Cross-lingual AUC at $N{=}10{,}000$ English
  training users. EN conditions are evaluated on JP real;
  JP Real uses inner CV on {\raise.17ex\hbox{$\scriptstyle\sim$}}4{,}000 labeled users
  (different protocol; reported as in-language reference, not directly comparable).
  $^{*}$\,$p{<}.01$, $^{**}$\,$p{<}.001$ (paired $z$-test vs Real EN). Overall spans all three activities; Paragraph (fixed content) shows gains are not from translated semantics alone.}
  \label{tab:rq2}
  \centering
  \resizebox{\columnwidth}{!}{%
  \begin{tabular}{lcccc}
    \toprule
    & \multicolumn{2}{c}{\textbf{Overall}} & \multicolumn{2}{c}{\textbf{Paragraph}} \\
    \cmidrule(lr){2-3} \cmidrule(lr){4-5}
    \textbf{Train condition} & \textbf{Depr.} & \textbf{Anx.} & \textbf{Depr.} & \textbf{Anx.} \\
    \midrule
    Real EN $\to$ JP Real & $0.593$ & $0.578$ & $0.594$ & $0.578$ \\
    \midrule
    CosyVoice 2 (EN$\to$JP) & $0.614^{**}$ & $0.609^{**}$ & $0.604^{**}$ & $\underline{0.608}^{**}$ \\
    CosyVoice 3 (EN$\to$JP) & $\underline{0.623}^{**}$ & $\mathbf{0.618}^{**}$ & $\mathbf{0.617}^{**}$ & $\mathbf{0.623}^{**}$ \\
    MaskGCT (EN$\to$JP) & $0.620^{**}$ & $\underline{0.612}^{**}$ & $0.605^{*}$ & $0.606^{**}$ \\
    OpenAudio (EN$\to$JP) & $\mathbf{0.626}^{**}$ & $0.605^{**}$ & $\underline{0.614}^{**}$ & $0.601^{**}$ \\
    \midrule
    JP Real (in-language) & $0.654$ & $0.664$ & $0.648$ & $0.676$ \\
    \bottomrule
  \end{tabular}}
\end{table}

\subsection{Scaling Analysis}
\label{sec:scaling}

\begin{figure}[t]
    \centering
    \includegraphics[width=\columnwidth]{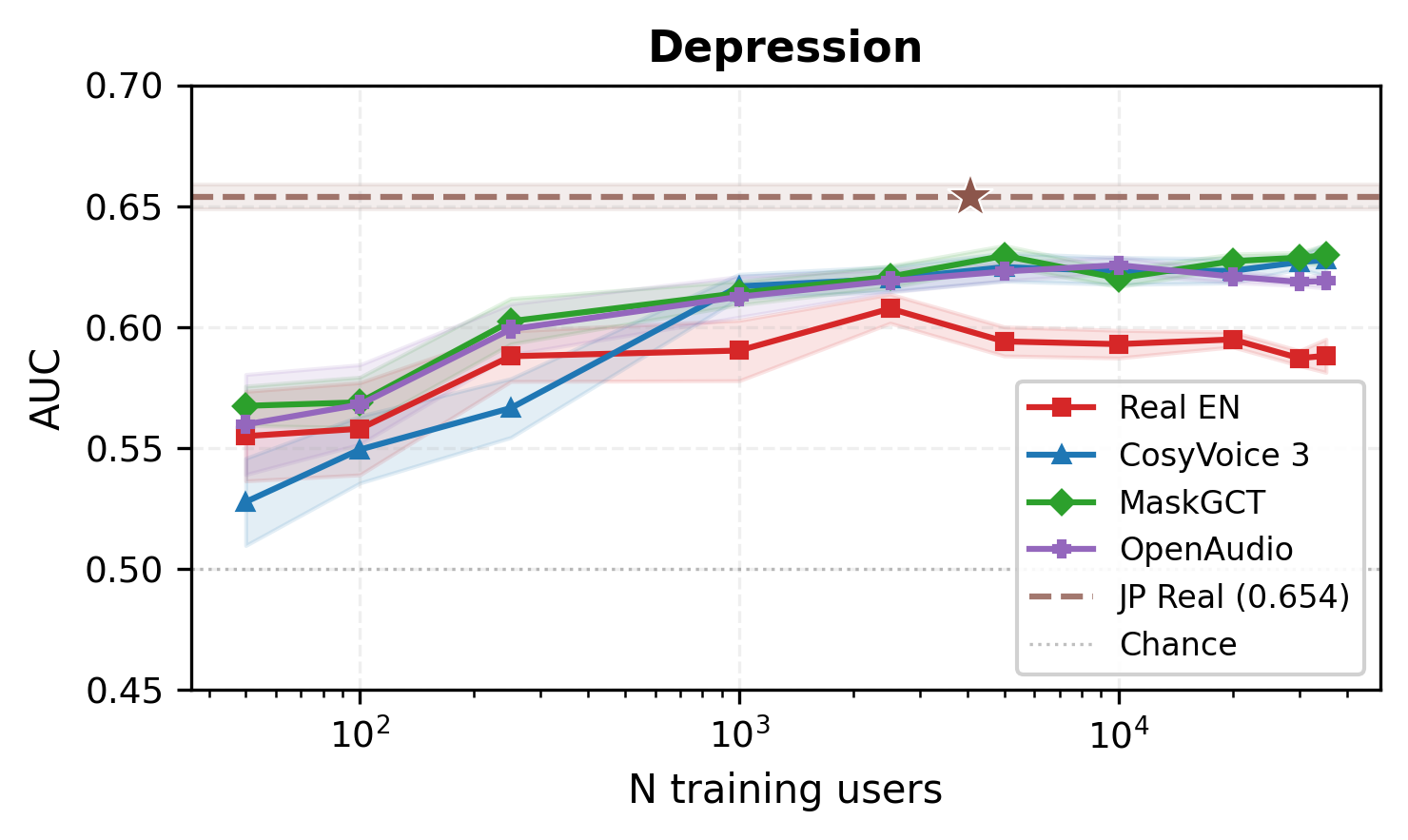}
    \caption{Scaling curves: AUC vs.\ number of training speakers for depression detection. Shaded bands show 95\% CI.}
    \label{fig:scaling}
\end{figure}

Figure~\ref{fig:scaling} plots depression AUC against the number of English training speakers for CosyVoice3, MaskGCT, OpenAudio and the real English baseline.
 Cloned models outperform the baseline from ${\sim}$1{,}000 speakers onward, rising steeply before plateauing, while the baseline shows no consistent gain beyond ${\sim}$2{,}500 speakers. Anxiety (omitted for space) follows a similar pattern, although with cloned models outperforming the baseline from the start ($N{=}50$). The in-language Japanese reference remains an upper bound that cloned models approach but do not reach.

\subsection{RQ3: Speaker similarity vs.\ downstream preservation} 
\begin{figure}[h]
    \centering
    \includegraphics[width=\columnwidth]{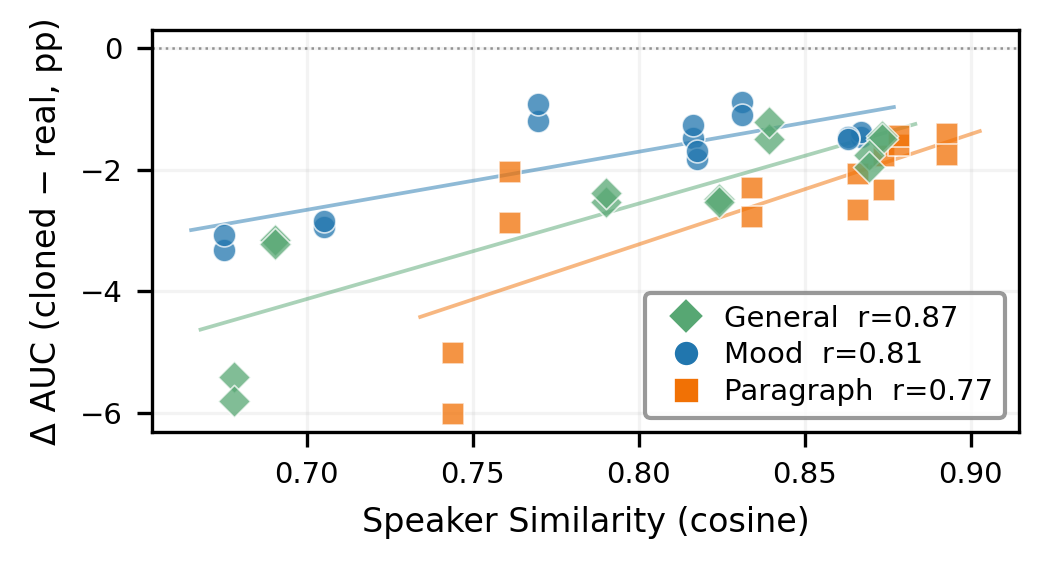}
    \caption{Speaker similarity vs.\ AUC degradation (percentage points) on the clinical corpus (repeat condition). Each point is one model; lines show per-activity linear fits.}
    \label{fig:similarity}
\end{figure}

Results show strong Pearson correlations between mean speaker similarity and AUC degradation on the clinical corpus (general $r{=}0.87$, mood $r{=}0.81$, paragraph $r{=}0.77$, Figure~\ref{fig:similarity}) and on IEMOCAP ($r{=}0.87$). 
Correlations are weaker for the TV-sourced MELD and MUSTARD ($r{=}0.18$, $r{=}0.30$), likely because background noise introduces degradation not
captured by speaker embeddings, and for VCTK ($r{=}0.57$). The trend is real but does not hold for all models: Zonos, for example, achieves high preservation on \textit{paragraph} despite moderate speaker similarity (an outlier that also drives down the VCTK correlation), suggesting some prosodic features are not fully reflected in embedding cosine distance.

\section{Discussion and Conclusion}
\label{sec:discussion}

Voice cloning is typically optimised for intelligibility and speaker similarity, not for preserving the signal that matters in paralinguistic tasks. Our results show that, despite this, modern cloning models retain most of the relevant signal. Under the repeat condition, the best models preserve over 90\% of the above-chance discriminative signal across all five tasks, including clinical depression and anxiety detection. 

Under the repeat condition, XTTS v2 and Zonos show the largest degradation, while flow-matching (E2-TTS, F5-TTS), hybrid (CosyVoice3), and masked generative (MaskGCT) architectures all perform better. However, the two weakest models have very different architectures (GPT-2 autoregressive vs. Mamba2 SSM), so architecture alone does not explain the difference.

The repeat versus standard comparison shows that models preserve different types of signal to different degrees. Under the standard condition, the ranking shifts: MaskGCT and E2-TTS retain the most signal (0.719 and 0.714 mean AUC), while OpenAudio, among the strongest under repeat, drops to the lowest (0.693). 
Accent is also transferred less faithfully when the generated text diverges from the reference audio (Section \ref{sec:results_rq1}).
Zonos shows a distinct pattern: relatively weak under repeat but third-strongest under standard (0.712), although this is largely due to its above-baseline accent performance and does not hold for other tasks. In practice, high repeat fidelity does not guarantee preservation of content-independent signal, and vice versa.

This preservation is robust enough to support cross-lingual augmentation. Training on English clinical speech cloned into Japanese consistently outperforms a raw English baseline, and the scaling analysis shows that this advantage emerges from as few as 1,000 source speakers, making augmentation practical even with modest source data. A gap remains to the in-language Japanese baseline, possibly due to language-specific differences in how depression and anxiety manifest prosodically, and because the translation and cloning steps may introduce additional noise.

Speaker similarity (RQ3) correlates well with downstream preservation on clean data but breaks down on noisy recordings, making it a useful but imperfect proxy. We recommend verifying preservation on a held-out subset of the target task before scaling augmentation.

\textbf{Limitations.} RQ2 covers only one language pair (English to Japanese); we have not tested whether this holds for other languages. The cross-lingual baseline is deliberately minimal, and stronger transfer methods may close the gap further. We use logistic regression throughout to control for classifier complexity, but results may differ with more expressive models. All features come from WavLM Large; other embeddings may yield different preservation patterns. We evaluate only open-source cloning models; proprietary systems may behave differently. Finally, the clinical results rely on a proprietary dataset, limiting external reproducibility.

\section{Use of Generative AI Disclosure}
Generative AI tools were used to assist with manuscript editing. All experimental results and scientific conclusions are solely the authors'.

\bibliographystyle{IEEEtran}
\bibliography{mybib}

\end{document}